\documentclass[runningheads]{llncs}

\usepackage[utf8]{inputenc}          
\usepackage[T1]{fontenc}             
\usepackage{lmodern}                 
\usepackage{microtype}               

\usepackage{amsmath,amssymb,mathtools}
\usepackage{bm}                       
\usepackage{nicefrac}                 

\usepackage{booktabs}                 
\usepackage{graphicx}                 
\graphicspath{{pictures/}{images/}{}} 
\DeclareGraphicsExtensions{.pdf,.png,.jpg}
\usepackage{multirow}
\usepackage{tabularx}
\usepackage{caption}
\usepackage{subcaption}
\usepackage{wrapfig}                  

\usepackage[dvipsnames]{xcolor}
\usepackage{color}                    

\usepackage[linesnumbered,ruled]{algorithm2e} 

\usepackage[numbers,square,sort&compress]{natbib}
\usepackage{doi}

\usepackage{lipsum}

\begin{document}

\title{Surrogate assisted diversity estimation in neural ensemble search\thanks{%
The final publication is available at Springer:
\url{https://link.springer.com/chapter/10.1007/978-3-032-30612-8_12}.\\
DOI: \url{https://doi.org/10.1007/978-3-032-30612-8_12}.
}}

\author{Alexandr Udeneev \and
Petr Babkin \and
Oleg Bakhteev}

\institute{
\email{\{udeneev.av, p.k.babkin, bakhteev.o\} (at) gmail.com}
}


\maketitle

\begin{abstract}
Ensembles are a standard way to improve the performance and robustness of deep neural networks, but their effectiveness crucially depends on both the quality and the diversity of individual models.
Most neural architecture search (NAS) methods are computationally expensive. Extending them to neural ensemble search (NES), which requires joint optimization of individual architectures and their ensemble composition, leads to an exponential growth of the search space and makes the problem computationally intractable. To address this, we introduce a dual-objective surrogate-guided ensemble search: candidate architectures are represented as directed acyclic graphs, and two surrogate models are trained independently to estimate predictive accuracy and diversity potential. Their combined estimates guide an NES framework that efficiently identifies architectures that are both individually strong and collectively diverse. Our final ensemble achieves competitive or superior performance compared to standard baselines such as Deep Ensembles and Random Search on FashionMNIST, CIFAR-10, and CIFAR-100.


\end{abstract}

\section{Introduction}

Neural network ensembles consistently demonstrate improved predictive accuracy, robustness, and uncertainty estimation compared to single models, particularly in classification and regression tasks \cite{Hansen1990, E_Ren_2016}. This observation motivates the study of \emph{Neural Ensemble Search} (NES), which aims to construct ensembles of neural networks in a principled and computationally efficient manner \cite{pmlr-v70-cortes17a, Zaidi2021}. In practice, NES relies on \emph{Neural Architecture Search} (NAS) techniques to identify suitable candidate architectures, a task that remains challenging due to the vastness of the search space and the high cost of training individual models \cite{Lu2022}.

A widely used baseline for ensemble construction is Deep Ensembles (DeepEns) \cite{lakshminarayanan2017simple}, which combine multiple independently trained instances of the same architecture. While simple and effective, this approach offers limited control over ensemble diversity, often resulting in highly correlated models. More recent NES methods adapt NAS techniques to ensemble construction \cite{Zaidi2021, Chen_2021_CVPR}, but diversity in these approaches typically emerges implicitly through optimization dynamics or heuristic similarity measures, rather than being explicitly modeled and optimized alongside accuracy.

\begin{figure}[h!]
    \centering
    \includegraphics[width=1\linewidth]{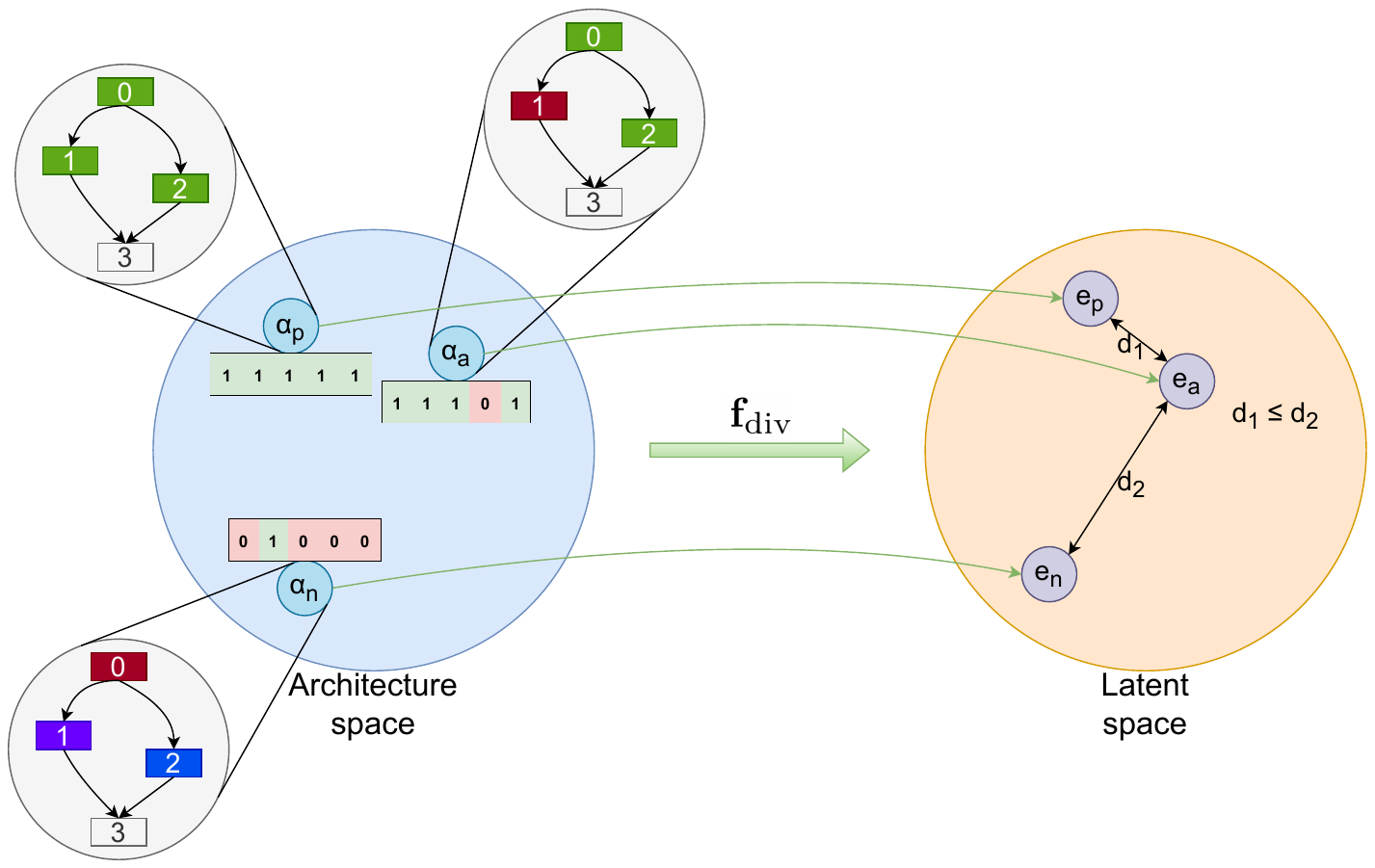}
    \caption{Illustration of the surrogate diversity function. Neural architectures are embedded into a latent space such that architectures with similar predictive behavior are mapped close together, while dissimilar ones are separated by larger distances. This enables efficient diversity estimation without training candidate models.}
    \label{fig:NES_surrogate}
\end{figure}

To address these limitations, we draw inspiration from surrogate-assisted NAS methods, which employ lightweight predictive models to estimate architecture performance without full training \cite{wen2020neural, white2021bananas}. While surrogates are commonly used to predict accuracy, recent work suggests that they can also capture structural and functional relationships between architectures \cite{S_Xue_2024}. This observation motivates their use for explicit diversity modeling in NES.

We propose a dual-objective surrogate-guided framework for Neural Ensemble Search. The framework separates the search into two complementary predictive objectives: accuracy estimation for individual architectures and diversity modeling via latent architecture embeddings. Candidate architectures are encoded as graphs and mapped into a continuous latent space using surrogate models, where geometric distances correspond to predictive dissimilarity. As illustrated in Fig.~\ref{fig:NES_surrogate}, this representation enables explicit, differentiable, and training-free evaluation of ensemble diversity. We implement the surrogate models using Graph Attention Networks (GATs) \cite{Velickovic2017} and structure the latent space using the triplet loss \cite{schroff2015facenet} for metric learning.

\vspace{2mm}

\textbf{Our main contributions are:}

\vspace{-2mm}

\begin{enumerate}
    \item We introduce a novel surrogate-based methodology for training diversity predictors that map discrete neural architectures to a structured latent space.
    \item We present the surrogate-assisted framework for Neural Ensemble Search that jointly optimizes predictive performance and architectural diversity via dual-objective guidance. To the best of our knowledge, this is one of the first methods to explicitly model diversity in NES using surrogate-based latent representations.
    \item We provide a proof-of-concept validation of our framework on FashionMNIST \cite{xiao2017fashion}, CIFAR-10 \cite{krizhevsky2009learning}, and CIFAR-100 \cite{krizhevsky2009learning}, demonstrating its capability to balance predictive performance and architectural diversity effectively.
    \item We provide datasets of approximately 3,000 trained models per dataset (including both weights and architectures) for FashionMNIST, CIFAR-10, and CIFAR-100~\cite{nes_dataset}.
    These datasets enable evaluation of ensemble methods without the need to train large numbers of models from scratch.
    Training details are provided in Section~\ref{section_4_1}.
\end{enumerate}

\section{Problem statement}
 Neural Architecture Search methods have proven effective at discovering high-performing individual models. However, many predictive tasks benefit from ensemble strategies that aggregate multiple models rather than relying on a single architecture. The challenge lies in the fact that an optimal ensemble requires not merely a collection of strong models, but a set of architectures that actively complement one another. This motivates our formulation of Neural Ensemble Search as a unified optimization problem, where architectural selection and ensemble composition are jointly optimized. We begin by reviewing the standard NAS framework, then extend it to the ensemble setting.

\subsection{Neural Architecture Search}
\label{section_2_1}

We follow the standard Neural Architecture Search (NAS) formulation~\cite{Liu2018}, where the goal is to identify an architecture $\bm{\alpha} \in \mathcal{A}$ that minimizes the validation loss after training:
\begin{equation}
    \begin{aligned}
        & \min_{\bm\alpha \in \mathcal{A}} \mathcal{L}_{val} (\bm\alpha, \bm\omega^*(\bm\alpha)), \\
        & \text{s.t.} \quad \bm\omega^*(\bm\alpha) = \arg \min_{\bm\omega \in \mathcal{W}} \mathcal{L}_{train} (\bm\alpha, \bm\omega).
    \end{aligned}
\end{equation}
Here, $\mathcal{A}$ denotes a search space of neural architectures represented as directed acyclic graphs with operations assigned to edges.

The primary challenge in NAS lies in the enormous size of the search space (e.g., $\sim 10^{25}$ in DARTS~\cite{Liu2018}), making exhaustive search infeasible.

\subsection{Neural Ensemble Search}

The primary objective of Neural Ensemble Search (NES) is to identify a set of complementary architectures $\mathcal{S} \subset \mathcal{A}$ that jointly minimize the validation loss.

We denote by $\bm{\alpha} \in \mathcal{A}$ a neural architecture and by $\bm{\omega}(\bm{\alpha})$ its corresponding parameters obtained by minimizing the training loss. The prediction of a model with architecture $\bm{\alpha}$ on input $\bm{x}$ is denoted by $f_{\bm{\alpha}}(\bm{x}, \bm{\omega}(\bm{\alpha}))$.

The NES problem can then be formulated as:
\begin{equation}
    \begin{aligned}
        & \min_{\mathcal{S} \subset \mathcal{A}} 
        \mathcal{L}_{val} \left( \frac{1}{|\mathcal{S}|} \sum_{\bm\alpha \in \mathcal{S}} f_{\bm\alpha}(\bm{x}, \bm\omega^*(\bm\alpha)) \right), \\
        & \text{s.t.} \quad \forall \bm\alpha \in \mathcal{S}: \ 
        \bm\omega^*(\bm\alpha) = \arg \min_{\bm\omega} \mathcal{L}_{train}(f_{\bm\alpha}(\bm{x}, \bm\omega)).
    \end{aligned}
\end{equation}

Compared to NAS, NES requires not only identifying high-performing individual architectures, but also selecting a subset of models that complement each other, significantly increasing the complexity of the search problem.

\section{Surrogate-based ensemble search}

In this work, we adapt the DARTS architecture space representation into a graph-structured form that can be processed by Graph Attention Networks (GATs) (see Section~\ref{section_3_1}). In Section~\ref{section_3_2}, we present the architecture of the surrogate functions and describe their operating principles. 
Section~\ref{section_3_3} then details the ensemble construction method based on these surrogate models, explaining how surrogate predictions are used to promote architectural diversity within the ensemble.

Together, these components enable the construction of ensembles that jointly consider predictive accuracy and architectural diversity.

\subsection{Architecture Search Space}\label{section_3_1}

Following the conventions of \cite{Liu2018}, we instantiate the search space $\mathcal{A}$ using a cell-based approach. Each architecture $\bm{\alpha} \in \mathcal{A}$ is composed of two functional units: a \textit{normal cell} $\bm{\alpha}_{\text{norm}}$ and a \textit{reduction cell} $\bm{\alpha}_{\text{red}}$, such that $\bm{\alpha} = (\bm{\alpha}_{\text{norm}}, \bm{\alpha}_{\text{red}})$. Both units are DAGs adhering to the formalisms established in Section \ref{section_2_1}.


Unlike the original DARTS, which relies on continuous relaxation, our method explores $\mathcal{A}$ via discrete random sampling. We consider a reduced configuration with $n=5$ nodes and $m=10$ edges, where each node has exactly two incoming and two outgoing edges. This choice follows the general structural design of DARTS, while reducing the size of the search space and the computational cost of training the surrogate dataset. Given an operation set size $|\mathcal{O}| = 7$, the cardinality of the cell-level search space is:
\begin{equation}
|\mathcal{A}_{\text{cell}}| = \prod_{k=2}^{n-1} \binom{k}{2} \cdot |\mathcal{O}|^m \approx 10^9.
\end{equation}
Consequently, the full architecture space $|\mathcal{A}| = |\mathcal{A}_{cell}|^2 \approx 10^{18}$ remains computationally intractable for exhaustive search, necessitating surrogate-guided exploration.

To train the dual-surrogate models, we construct a dataset $\mathcal{D}_{\text{train}}$ by evaluating $N$ sampled architectures:
\begin{equation}
\mathcal{D}_{\text{train}} = \{(\bm{\alpha}_i, \mathbf{y}_i, \text{acc}_i)\}_{i=1}^{N},
\end{equation}
where $\mathbf{y}_i = f_{\bm{\alpha}_i}(\mathcal{X}_{val}, \bm{\omega}^*(\bm{\alpha}_i))$ is the vector of model predictions on a fixed validation subset $\mathcal{X}_{val}$, and $\text{acc}_i$ is the corresponding validation accuracy. These observations serve as the ground truth for our surrogate functions.

\subsection{Surrogate Function}  
\label{section_3_2}

To facilitate ensemble construction (Section~\ref{section_3_3}), we estimate both the performance and diversity of candidate architectures using surrogate models. Due to the size of the search space, these properties cannot be obtained via full training. We therefore introduce two surrogate functions:
\begin{equation}
  f_{\mathrm{acc}}^{\bm{\theta}}: \mathcal{A} \to \mathbb{R}, \quad
  \mathbf{f}_{\mathrm{div}}^{\bm{\theta}}: \mathcal{A} \to \mathbb{R}^d.
\end{equation}

Each architecture is represented as a directed acyclic graph. For surrogate modeling, we use an operation-centric graph representation, where each edge of the original architecture is treated as a node, and connections are introduced between sequential operations. Operations are represented via one-hot encoding.

Both surrogates are implemented as Graph Attention Networks (GATs) with residual connections, GraphNorm~\cite{cai2021graphnorm}, and global pooling.

The accuracy surrogate $f_{\mathrm{acc}}$ is trained via supervised regression using the mean squared error:
\begin{equation}
    \bm{\theta}^*_{\text{acc}} = \arg\min_{\bm{\theta}} \sum_{i=1}^{N} \left( f^{\bm{\theta}}_{\text{acc}}(\bm{\alpha}_i) - \text{acc}_i \right)^2.
\end{equation}

The diversity surrogate $\mathbf{f}_{\mathrm{div}}$ is trained using a triplet loss to enforce proximity between architectures with similar predictions and separation otherwise. Positive and negative pairs are defined based on high and low agreement in model predictions.

Given model predictions $\{\bm{y}^{(i)}\}_{i=1}^N$ on a validation set, we define a similarity matrix:
$$
c_{ij} = \frac{1}{K} \sum_{k=1}^{K} \mathbb{I}\left( y^{(i)}_k = y^{(j)}_k \right),
$$
which is discretized into $\mathbf{D} \in \{-1,0,1\}^{N \times N}$ using quantile thresholds.

The surrogate maps architectures into a latent space:
\begin{equation}
    \mathbf{e}_a = \mathbf{f}^{\bm{\theta}_{\mathrm{div}}}_{\mathrm{div}}(\alpha_a),
\end{equation}
and is optimized via the triplet loss:
\begin{equation}
    \mathcal{L}_{\mathrm{div}} = \sum_{(a,p,n)} \max\Big(
    \|\mathbf{e}_a - \mathbf{e}_p\|_2^2 - \|\mathbf{e}_a - \mathbf{e}_n\|_2^2 + m, 0
    \Big).
\end{equation}

This training encourages a latent space where Euclidean distances reflect predictive diversity. The full training procedure is summarized in Fig.~\ref{fig:surrogate_training}.

\begin{figure}[h!]
\centering

\setcounter{AlgoLine}{0}
\begin{algorithm}[H]
\SetAlgoLined
\DontPrintSemicolon
\KwIn{
    $\mathcal{D}_{train}$: training dataset of architectures $\{(\bm{\alpha}_i, \mathbf{y}_i, \text{acc}_i)\}_{i=1}^N$; \\
    $N$: number of architectures in the dataset; \\
    $n$: number of training epochs; \\
    $\mathbf{D}$: diversity matrix where $\mathbf{D}_{jk} \in \{-1, 0, 1\}$ (negative/neutral/positive); \\
    $m$: triplet loss margin; \\
    $\eta$: learning rate; \\
    $B$: batch size.
}
\KwOut{Optimized parameters $\bm{\theta}_{div}$ for surrogate model $\mathbf{f}_{div}$}

Initialize parameters $\bm{\theta}_{div}$\;

\For{epoch $e \gets 1$ \KwTo $n$}{
    \For{step $t \gets 1$ \KwTo $N/B$}{
        Sample a minibatch of anchor indices $\mathcal{J} \subset \{1, \dots, N\}$, where $|\mathcal{J}| = B$\;
        
        \For{each $j \in \mathcal{J}$}{
            Identify candidate sets: $\mathcal{P}_j = \{ k \mid \mathbf{D}_{jk} = 1 \}$ and $\mathcal{N}_j = \{ k \mid \mathbf{D}_{jk} = -1 \}$\;
            Sample positive $k_p \sim \text{Uniform}(\mathcal{P}_j)$ and negative $k_n \sim \text{Uniform}(\mathcal{N}_j)$\;
            
            Compute embeddings: \\
            $\mathbf{e}_a, \mathbf{e}_p, \mathbf{e}_n \gets \mathbf{f}^{\bm{\theta}_{div}}_{div}(\bm{\alpha}_j), \mathbf{f}^{\bm{\theta}_{div}}_{div}(\bm{\alpha}_{k_p}), \mathbf{f}^{\bm{\theta}_{div}}_{div}(\bm{\alpha}_{k_n})$\;
            
            Compute triplet loss for instance $j$: \\
            $\ell_j \gets \max\left(0, || \mathbf{e}_a - \mathbf{e}_p ||_2^2 - || \mathbf{e}_a - \mathbf{e}_n ||_2^2 + m \right)$\;
        }
        
        Compute batch loss: $\mathcal{L} = \frac{1}{B} \sum_{j \in \mathcal{J}} \ell_j$\;
        Update parameters: $\bm{\theta}_{div} \gets \bm{\theta}_{div} - \eta \nabla_{\bm{\theta}_{div}} \mathcal{L}$
    }
}
\Return{$\bm{\theta}_{div}$}
\end{algorithm}

\caption{Training procedure for the diversity surrogate function.}
\label{fig:surrogate_training}

\end{figure}

\subsection{Ensemble Construction}\label{section_3_3}

Once the surrogate models $f^{\bm{\theta}_{\mathrm{acc}}}_{\text{acc}}$ and $\mathbf{f}^{\bm{\theta}_{\mathrm{div}}}_{\text{div}}$ are trained, they enable an efficient search for the optimal ensemble without requiring the prohibitive computational cost of training intermediate candidate architectures. The proposed ensemble construction (Fig~\ref{fig:greedy_ensemble}) proceeds in two distinct phases: candidate pool filtering and diversity-driven greedy selection.

\begin{figure}[h!]
\centering

\begin{algorithm}[H]
\SetAlgoLined
\DontPrintSemicolon
\KwIn{
    $K$: target ensemble size; \\
    $N$: number of initial candidates; \\
    $\lambda$: accuracy threshold; \\
    $f_{\text{acc}}, \mathbf{f}_{\text{div}}$: trained surrogate models.
}
\KwOut{
    $\mathcal{A}_{\text{best}}$: selected ensemble of $K$ architectures.
}

\tcp{Phase 1: Candidate Pool Construction}
$\mathcal{A}_{\text{cand}} \gets \text{RandomSample}(N)$\;
$\mathcal{P} \gets \bigl\{ (\alpha, a, \mathbf{e}) \mid \alpha \in \mathcal{A}_{\text{cand}}, a = f_{\text{acc}}(\alpha), \mathbf{e} = \mathbf{f}_{\text{div}}(\alpha), a \ge \lambda \bigr\}$\;

\tcp{Phase 2: Diversity-driven Greedy Selection}
Select triplet with highest predicted accuracy: $(\alpha^*, a^*, \bm{e}^*) = \arg\max_{(\alpha, a, \mathbf{e})\in \mathcal{P}} a$\;
$\mathcal{A}_{\text{best}} \gets \{\alpha^*\}$; \quad $\mathcal{E}_{\text{sel}} \gets \{\mathbf{e}^*\}$; \quad $\mathcal{P} \gets \mathcal{P} \setminus \{(\alpha^*, a^*, \mathbf{e}^*)\}$\;

\While{$|\mathcal{A}_{\text{best}}| < K$ \textbf{and} $\mathcal{P} \neq \emptyset$}{
    For each candidate $(\alpha_i, a_i, \mathbf{e}_i) \in \mathcal{P}$, compute
    $d_i = \frac{1}{|\mathcal{E}_{\text{sel}}|} \sum_{\mathbf{e}_{\text{sel}} \in \mathcal{E}_{\text{sel}}} \| \mathbf{e}_i - \mathbf{e}_{\text{sel}} \|_2$\;
    
    Select triplet with maximum diversity: $(\alpha_{i^*}, a_{i^*}, \mathbf{e}_{i^*}) = \arg\max_{(\alpha_i, a_i, \mathbf{e}_i) \in \mathcal{P}} d_i$\;
    Update ensemble: $\mathcal{A}_{\text{best}} \gets \mathcal{A}_{\text{best}} \cup \{\alpha_{i^*}\}$; \quad $\mathcal{E}_{\text{sel}} \gets \mathcal{E}_{\text{sel}} \cup \{\mathbf{e}_{i^*}\}$\;
    $\mathcal{P} \gets \mathcal{P} \setminus \{(\alpha_{i^*}, a_{i^*}, \mathbf{e}_{i^*})\}$\;
}

\Return{$\mathcal{A}_{\text{best}}$}
\end{algorithm}

\caption{Surrogate-Assisted Ensemble Construction.}
\label{fig:greedy_ensemble}

\end{figure}

In the first phase, we generate a large set of random candidate architectures $\mathcal{A}_{\text{cand}}$ via discrete sampling from the search space. We then use the dual-surrogate framework to predict the accuracy $a$ and the latent diversity embedding $\mathbf{e}$ for each candidate. To ensure the base quality of the ensemble, we apply an accuracy threshold $\lambda$, forming a pruned candidate pool $\mathcal{P}$ that contains only high-performing models.

The second phase employs a greedy forward-selection strategy to maximize architectural diversity. Starting with the most accurate model as an anchor, the algorithm iteratively selects subsequent architectures that exhibit the maximum average Euclidean distance from the currently selected set in the latent space. By leveraging the geometric organization of the latent space provided by $\mathbf{f}_{\text{div}}$, this selection process explicitly favors architectures with dissimilar prediction patterns. This approach effectively identifies a diverse set of high-performing models, balancing individual predictive power with collective ensemble robustness.

\section{Computational Experiments}

In this section, we present the experimental results as well as the metrics used for comparison. In Section \ref{section_4_1}, we describe the dataset collection procedure. Section \ref{section_4_2} details the training setup for the surrogate functions. Section \ref{sec:ablations_surrogate} presents ablation studies analyzing the contribution of each surrogate function and the number of trained models required for surrogate learning. Finally, in Section \ref{sec:ensemble_results}, we compare our proposed method against DeepEns  \cite{lakshminarayanan2017simple} and Random Search \cite{li2020random}  for FashionMNIST, CIFAR-10 and CIFAR-100, respectively.

\subsection{Construction of the Training Dataset for Surrogate Models}
\label{section_4_1}

The models in our dataset are trained according to the process described in Fig.~\ref{fig:dataset_construction}. Training follows \cite{Liu2018} with minor modifications; the main hyperparameters are summarized in Table~\ref{tab:training-params-dataset}.

\begin{table}[h]
    \centering
    \caption{Training Hyperparameters and Performance per Dataset}
    \label{tab:training-params-dataset}
    \begin{tabularx}{\textwidth}{l|c|c|c|c|X}
        \hline
        Dataset &
        Num.~Cells &
        Initial~Width &
        Num.~Epochs &
        Avg.~Acc.~(\%) &
        Avg. Top-1 ~Agreement \\
        \hline
        FashionMNIST & 3 & 16 & 125 & $89.6 \pm 0.5$ & $0.900 \pm 0.004$ \\
        CIFAR-10     & 8 & 16 & 200 & $75.8 \pm 0.6$ & $0.693 \pm 0.006$ \\
        CIFAR-100    & 8 & 16 & 200 & $37.6 \pm 1.1$ & $0.324 \pm 0.008$ \\
        \hline
    \end{tabularx}
\end{table}

It is important to note that the architectures used to construct the surrogate training dataset are intentionally trained in a reduced configuration (in terms of depth and width) to limit computational cost. Moreover, these models are not trained to full convergence, but only for a fixed number of epochs sufficient to obtain reliable relative performance estimates.

In contrast, the final ensemble models are trained using a larger configuration, with increased number of cells and channel width. This design allows the surrogate functions, learned on lightweight and partially trained models, to effectively guide the selection of higher-capacity architectures during ensemble construction, significantly reducing the overall computational cost.

\begin{figure}[h!]
    \centering
    \begin{algorithm}[H]
    \SetAlgoLined
    \DontPrintSemicolon
    \KwIn{
        $\mathcal{A}_{DARTS}$: architectures available through DARTS; \\
        $\mathcal{X}$: dataset for train and validate architectures; \\
        $N$: number of architectures.
    }
    \KwOut{
        Model dataset 
        $
        \mathcal{D} = \{(\bm{\alpha}_i, \bm{y}_i, \text{acc}_i)\}_{i=1}^{N}.
        $
    }

    Split dataset $\mathcal{X}$ into training and validation subsets:
    $\mathcal{X}_{\text{train}}, \mathcal{X}_{\text{val}}$ with a $20\%/80\%$ ratio\;

    \For{$i \gets 1$ \KwTo $N$}{
        Sample architecture 
        $
        \bm{\alpha}_i \sim \text{Uniform}(\mathcal{A}_{DARTS})
        $\;

        Train model 
        $
        f_{\bm{\alpha}_i}(\mathcal{X}_{\text{train}}, \bm{\omega}(\bm{\alpha}_i))
        $\;

        Evaluate $f_{\bm{\alpha}_i}$ on $\mathcal{X}_{\text{val}}$\;
        \Indp
            prediction vector
            $
            \bm{y}_i = f_{\bm{\alpha}_i}(\mathcal{X}_{\text{val}}, \bm{\omega}^*(\bm{\alpha}_i))
            $\;

            validation accuracy
            $
            \text{acc}_i = \frac{1}{|\mathcal{X}_{\text{val}}|} 
            \sum_{(x,y)\in \mathcal{X}_{\text{val}}} 
            \mathbb{I}\big(
            \arg\max_c f_{\bm{\alpha}_i}(x;\bm{\omega}^*(\bm{\alpha}_i)) = y
            \big)
            $\;

            $\mathcal{D} \leftarrow \mathcal{D} \cup (\bm{\alpha}_i, \bm{y}_i, \text{acc}_i)$\;
        \Indm
    }

    \Return{$\mathcal{D}$}
    \end{algorithm}
    \caption{Construction of the Model Evaluation Dataset}
    \label{fig:dataset_construction}
\end{figure}

To improve efficiency, we employ an iterative surrogate-assisted refinement: we first train the accuracy surrogate on $N_1 = 1000$ randomly sampled architectures, then restrict the search to the top $10\%$ candidates and sample an additional $N_2 = 2000$ architectures from this reduced space.

Constructing the surrogate dataset remains the main computational bottleneck. 
In our experiments, training approximately 3,000 architectures for each dataset, CIFAR-10 and CIFAR-100, required about three days per dataset on four NVIDIA A100 GPUs (80 GB each).

\subsection{Training of the Surrogate Functions}
\label{section_4_2}

Each architecture is represented as a cell with $n = 5$ nodes and $m = 10$ edges, where operations are sampled from a standard DARTS search space, including separable and dilated convolutions, pooling, and skip connections.

We use a Graph Attention Network (GAT) with four convolutional layers and two fully connected layers to model both surrogate functions, trained with a cosine annealing learning rate schedule.

Full models are constructed by stacking normal and reduction cells in a $2{:}1$ ratio with predefined initial channels.

\subsection{Ablations of the Surrogate Functions}
\label{sec:ablations_surrogate}

The main computational bottleneck of the proposed framework is the need to construct a large dataset of trained architectures. In this section, we analyze how the performance of the surrogate models depends on the training set size. All experiments are conducted on CIFAR-100 as the most challenging dataset.

We begin with the accuracy surrogate, which is used to filter candidate architectures based on their predicted performance. We hold out 600 architectures and evaluate how well the surrogate retrieves high-performing models using Spearman correlation and Recall@K.

The results are presented in Table~\ref{tab:surrogate_ablation}. Both Spearman correlation and Recall@K improve as the number of training architectures increases, indicating that the surrogate becomes more effective at identifying strong models.

\begin{table}[h!]
\centering
\caption{Surrogate performance as a function of training set size. Results are averaged over 10 runs; subscripts denote standard deviation.}
\label{tab:surrogate_ablation}
\begin{tabular}{c|c|c|ccc}
\toprule
\textbf{Type} & \textbf{Train Size} & \textbf{Spearman} & \textbf{R@10} & \textbf{R@50} & \textbf{R@100} \\
\midrule
\multirow{7}{*}{Accuracy}
& 0    & $0.017_{\pm 0.051}$ & $0.010_{\pm 0.030}$ & $0.076_{\pm 0.017}$ & $0.167_{\pm 0.021}$ \\
& 250  & $0.106_{\pm 0.106}$ & $0.040_{\pm 0.049}$ & $0.130_{\pm 0.039}$ & $0.228_{\pm 0.052}$ \\
& 500  & $0.220_{\pm 0.058}$ & $0.090_{\pm 0.083}$ & $0.196_{\pm 0.064}$ & $0.291_{\pm 0.034}$ \\
& 750  & $0.218_{\pm 0.079}$ & $0.120_{\pm 0.125}$ & $0.204_{\pm 0.061}$ & $0.291_{\pm 0.046}$ \\
& 1000 & $0.368_{\pm 0.050}$ & $0.120_{\pm 0.087}$ & $0.264_{\pm 0.073}$ & $0.330_{\pm 0.025}$ \\
& 1500 & $0.535_{\pm 0.060}$ & $0.170_{\pm 0.135}$ & $0.294_{\pm 0.073}$ & $0.366_{\pm 0.032}$ \\
& 2000 & $0.675_{\pm 0.021}$ & $0.200_{\pm 0.089}$ & $0.324_{\pm 0.032}$ & $0.430_{\pm 0.031}$ \\
\midrule
\multirow{7}{*}{Diversity}
& 0    & $-0.001_{\pm 0.029}$ & $0.019_{\pm 0.003}$ & $0.092_{\pm 0.005}$ & $0.179_{\pm 0.006}$ \\
& 250  & $-0.490_{\pm 0.042}$ & $0.029_{\pm 0.004}$ & $0.131_{\pm 0.008}$ & $0.249_{\pm 0.011}$ \\
& 500  & $-0.545_{\pm 0.035}$ & $0.031_{\pm 0.003}$ & $0.136_{\pm 0.005}$ & $0.261_{\pm 0.008}$ \\
& 750  & $-0.574_{\pm 0.025}$ & $0.029_{\pm 0.003}$ & $0.135_{\pm 0.006}$ & $0.259_{\pm 0.010}$ \\
& 1000 & $-0.602_{\pm 0.025}$ & $0.031_{\pm 0.003}$ & $0.142_{\pm 0.007}$ & $0.271_{\pm 0.012}$ \\
& 1500 & $-0.639_{\pm 0.022}$ & $0.032_{\pm 0.003}$ & $0.147_{\pm 0.005}$ & $0.281_{\pm 0.007}$ \\
& 2000 & $-0.637_{\pm 0.025}$ & $0.032_{\pm 0.004}$ & $0.145_{\pm 0.007}$ & $0.278_{\pm 0.008}$ \\
\bottomrule
\end{tabular}
\end{table}

The diversity surrogate is the core component of our method. To evaluate it, we again hold out 600 architectures. We measure (i) Spearman correlation between distances in the latent space and prediction similarity, and (ii) Recall@K, which reflects how well similar models are placed close to each other in the embedding space.

For the diversity surrogate, the Spearman correlation is negative, as larger distances in the latent space correspond to lower similarity between model predictions.

We observe that the diversity surrogate rapidly learns a meaningful latent structure: both correlation and Recall@K improve significantly when increasing the number of training architectures up to approximately 500--1000 samples, after which the gains become marginal. This suggests that reliable diversity estimation can be achieved without requiring the full surrogate training dataset.

We additionally perform a component ablation to assess the contribution of each surrogate function. The results are presented in Table~\ref{tab:component_ablation}.

\begin{table}[h!]
\centering
\caption{Component ablation on CIFAR-100.}
\label{tab:component_ablation}
\begin{tabular}{l|ccc}
\toprule
\textbf{Method} & \textbf{Top-1 Acc. (\%)} & \textbf{Avg. Model Acc. (\%)} & \textbf{Pred. Disagreement} \\
\midrule
Random & $84.41_{\pm 0.09}$ & $79.57_{\pm 0.19}$ & $\mathbf{0.430_{\pm 0.004}}$ \\
Accuracy only & $85.07_{\pm 0.10}$ & $\mathbf{80.54_{\pm 0.12}}$ & $0.406_{\pm 0.002}$ \\
Diversity only & $84.77_{\pm 0.07}$ & $80.08_{\pm 0.21}$ & $0.419_{\pm 0.008}$ \\
Accuracy + Diversity & $\mathbf{85.17_{\pm 0.16}}$ & $80.50_{\pm 0.12}$ & $0.411_{\pm 0.004}$ \\
\bottomrule
\end{tabular}
\end{table}

\subsection{Ensemble Performance Across Datasets}
\label{sec:ensemble_results}

Across all datasets, the proposed Surrogate Ensemble consistently achieves the best or competitive performance in terms of Top-1 accuracy and NLL (Table~\ref{tab:ensemble_results}). The DeepEns baseline is constructed by repeatedly training a single architecture obtained via DARTS with different random initializations.

On FashionMNIST, all methods perform similarly due to the low task complexity; however, the Surrogate Ensemble attains the highest accuracy and average model performance. While Random Search yields higher diversity, this does not translate into improved ensemble accuracy.

\begin{table*}[h!]
\centering
\caption{Ensemble performance on FashionMNIST, CIFAR-10, and CIFAR-100.}
\label{tab:ensemble_results}
\begin{tabular}{clccc}
\toprule
\textbf{Dataset} & \textbf{Metric} & \textbf{Surrogate Ens.} & \textbf{DeepEns} & \textbf{Random Search} \\
\midrule

\multirow{11}{*}{\rotatebox[origin=c]{90}{\textbf{FashionMNIST}}}
& Top-1 Acc. (\%) & $\mathbf{95.3_{\pm 0.1}}$ & $95.02_{\pm 0.35}$ & $95.01_{\pm 0.19}$ \\
& Avg. Model Acc. (\%) & $\mathbf{94.7_{\pm 0.1}}$ & $94.55_{\pm 0.43}$ & $94.32_{\pm 0.24}$ \\
& NLL & $\mathbf{0.263_{\pm 0.003}}$ & $0.266_{\pm 0.006}$ & $0.265_{\pm 0.003}$ \\
& Oracle NLL & $0.199_{\pm 0.010}$ & $0.206_{\pm 0.003}$ & $\mathbf{0.191_{\pm 0.010}}$ \\
& Brier Score & $\mathbf{0.089_{\pm 0.001}}$ & $0.091_{\pm 0.006}$ & $0.092_{\pm 0.002}$ \\
& ECE & $0.124_{\pm 0.002}$ & $0.121_{\pm 0.001}$ & $\mathbf{0.120_{\pm 0.004}}$ \\
& Ambiguity & $0.0058_{\pm 0.0009}$ & $0.0047_{\pm 0.0009}$ & $\mathbf{0.0069_{\pm 0.0013}}$ \\
& Norm. Disagreement & $0.065_{\pm 0.008}$ & $0.062_{\pm 0.008}$ & $\mathbf{0.077_{\pm 0.009}}$ \\
& Pred. Disagreement & $0.130_{\pm 0.016}$ & $0.125_{\pm 0.016}$ & $\mathbf{0.154_{\pm 0.018}}$ \\
& FGSM AUC & $\mathbf{0.058_{\pm 0.0025}}$ & $0.056_{\pm 0.0036}$ & $0.056_{\pm 0.0012}$ \\
& PGD AUC & $0.023_{\pm 0.0018}$ & $\mathbf{0.024_{\pm 0.0029}}$ & $0.022_{\pm 0.0010}$ \\

\midrule
\multirow{11}{*}{\rotatebox[origin=c]{90}{\textbf{CIFAR-10}}}
& Top-1 Acc. (\%) & $\mathbf{97.80_{\pm 0.08}}$ & $97.64_{\pm 0.09}$ & $97.56_{\pm 0.11}$ \\
& Avg. Model Acc. (\%) & $\mathbf{96.73_{\pm 0.10}}$ & $96.71_{\pm 0.14}$ & $96.48_{\pm 0.09}$ \\
& NLL & $\mathbf{0.208_{\pm 0.003}}$ & $0.210_{\pm 0.003}$ & $0.212_{\pm 0.002}$ \\
& Oracle NLL & $0.156_{\pm 0.002}$ & $\mathbf{0.154_{\pm 0.004}}$ & $0.158_{\pm 0.003}$ \\
& Brier Score & $\mathbf{0.055_{\pm 0.001}}$ & $0.056_{\pm 0.001}$ & $0.057_{\pm 0.001}$ \\
& ECE & $0.136_{\pm 0.002}$ & $\mathbf{0.135_{\pm 0.002}}$ & $0.136_{\pm 0.001}$ \\
& Ambiguity & $\mathbf{0.011_{\pm 0.001}}$ & $0.009_{\pm 0.001}$ & $\mathbf{0.011_{\pm 0.001}}$ \\
& Norm. Disagreement & $0.044_{\pm 0.001}$ & $0.046_{\pm 0.003}$ & $\mathbf{0.046_{\pm 0.001}}$ \\
& Pred. Disagreement & $0.087_{\pm 0.003}$ & $\mathbf{0.093_{\pm 0.007}}$ & $0.092_{\pm 0.003}$ \\
& FGSM AUC & $\mathbf{0.0569_{\pm 0.0012}}$ & $0.0555_{\pm 0.0021}$ & $0.0559_{\pm 0.0012}$ \\
& PGD AUC & $\mathbf{0.0273_{\pm 0.0015}}$ & $0.0251_{\pm 0.0020}$ & $0.0256_{\pm 0.0018}$ \\

\midrule
\multirow{11}{*}{\rotatebox[origin=c]{90}{\textbf{CIFAR-100}}}
& Top-1 Acc. (\%) & $\mathbf{85.17_{\pm 0.16}}$ & $85.10_{\pm 0.29}$ & $84.41_{\pm 0.09}$ \\
& Avg. Model Acc. (\%) & $80.50_{\pm 0.12}$ & $\mathbf{80.97_{\pm 0.41}}$ & $79.57_{\pm 0.19}$ \\
& NLL & $0.692_{\pm 0.007}$ & $\mathbf{0.682_{\pm 0.010}}$ & $0.718_{\pm 0.006}$ \\
& Oracle NLL & $\mathbf{0.403_{\pm 0.004}}$ & $0.406_{\pm 0.007}$ & $0.418_{\pm 0.006}$ \\
& Brier Score & $0.234_{\pm 0.002}$ & $\mathbf{0.232_{\pm 0.004}}$ & $0.244_{\pm 0.002}$ \\
& ECE & $0.140_{\pm 0.004}$ & $\mathbf{0.133_{\pm 0.003}}$ & $0.141_{\pm 0.002}$ \\
& Ambiguity & $0.047_{\pm 0.001}$ & $0.041_{\pm 0.002}$ & $\mathbf{0.048_{\pm 0.002}}$ \\
& Norm. Disagreement & $0.206_{\pm 0.002}$ & $0.195_{\pm 0.005}$ & $\mathbf{0.215_{\pm 0.002}}$ \\
& Pred. Disagreement & $0.411_{\pm 0.004}$ & $0.390_{\pm 0.010}$ & $\mathbf{0.430_{\pm 0.004}}$ \\
& FGSM AUC & $\mathbf{0.037_{\pm 0.001}}$ & $0.0368_{\pm 0.001}$ & $0.0358_{\pm 0.001}$ \\
& PGD AUC & $\mathbf{0.017_{\pm 0.001}}$ & $0.0158_{\pm 0.001}$ & $0.0167_{\pm 0.0005}$ \\

\bottomrule
\end{tabular}
\end{table*}

On CIFAR-10, the advantage of the proposed approach becomes more evident. The Surrogate Ensemble achieves the best Top-1 accuracy and NLL, while maintaining competitive diversity. This indicates that the method effectively balances individual model quality and ensemble diversity.

On CIFAR-100, the most challenging dataset, the Surrogate Ensemble achieves the highest Top-1 accuracy despite slightly lower average individual model accuracy compared to DeepEns. This highlights the benefit of selecting complementary architectures rather than optimizing models independently. In contrast, Random Search exhibits higher diversity but suffers from weaker individual models, leading to inferior ensemble performance.

Overall, these results demonstrate that explicitly modeling diversity is beneficial for constructing effective ensembles, particularly as dataset complexity increases.

\section{Conclusion}

In this study, we showed that the proposed surrogate diversity function learns a latent embedding of model architectures in which inter-model distances consistently reflect high levels of predictive diversity across FashionMNIST, CIFAR-10, and CIFAR-100. Building on this observation, we proposed a surrogate-guided ensemble construction framework that jointly optimizes accuracy and diversity, demonstrating consistent improvements over standard baselines. Our results further indicate that the role of diversity becomes increasingly important as dataset complexity grows, highlighting the necessity of principled diversity modeling in challenging settings.

A key limitation of the proposed approach is the computational cost associated with training a large pool of base models to construct the surrogate dataset. Future work may mitigate this overhead through more efficient sampling strategies or weight-sharing mechanisms, as well as by replacing discrete similarity measures with continuous information-theoretic metrics to improve the smoothness and expressiveness of diversity estimation.

\bibliographystyle{unsrtnat}
\renewcommand{\bibname}{References}
\bibliography{refs}

@article{E_Ren_2016,
  title   = {Ensemble Classification and Regression--Recent Developments, Applications and Future Directions},
  author  = {Ren, Ye and Zhang, Le and Suganthan, P. N.},
  year    = {2016},
  month   = feb,
  journal = {IEEE Computational Intelligence Magazine},
  volume  = {11},
  number  = {1},
  pages   = {41--53},
  doi     = {10.1109/MCI.2015.2471235},
  issn    = {1556-603X}
}

@inproceedings{Zaidi2021,
  author    = {Zaidi, Sheheryar and Zela, Arber and Elsken, Thomas and Holmes, Chris C. and Hutter, Frank and Teh, Yee Whye},
  title     = {Neural Ensemble Search for Uncertainty Estimation and Dataset Shift},
  booktitle = {Advances in Neural Information Processing Systems 34},
  year      = {2021},
  pages     = {7898--7911},
  editor    = {M. Ranzato and A. Beygelzimer and Y. Dauphin and P.S. Liang and J. Wortman Vaughan}
}

@article{Hansen1990,
  author  = {Hansen, Lars Kai and Salamon, Peter},
  title   = {Neural Network Ensembles},
  journal = {IEEE Transactions on Pattern Analysis and Machine Intelligence},
  year    = {1990},
  volume  = {12},
  number  = {10},
  pages   = {993--1001},
  doi     = {10.1109/34.58871}
}

@article{Lu2022,
  author        = {Lu, Zhichao and Cheng, Ran and Huang, Shihua and Zhang, Haoming and Qiu, Changxiao and Yang, Fan},
  title         = {Surrogate-assisted Multi-objective Neural Architecture Search for Real-time Semantic Segmentation},
  journal       = {CoRR},
  volume        = {abs/2208.06820},
  year          = {2022},
  eprint        = {2208.06820},
  archivePrefix = {arXiv},
  doi           = {10.48550/ARXIV.2208.06820}
}

@article{S_Xue_2024,
  title   = {Similarity surrogate-assisted evolutionary neural architecture search with dual encoding strategy},
  author  = {Xue, Yu and Zhang, Zhenman and Neri, Ferrante},
  year    = {2024},
  journal = {Electronic Research Archive},
  volume  = {32},
  number  = {2},
  pages   = {1017--1043},
  doi     = {10.3934/era.2024050},
  issn    = {2688-1594}
}

@inproceedings{Liu2018,
  author        = {Liu, Hanxiao and Simonyan, Karen and Yang, Yiming},
  title         = {DARTS: Differentiable Architecture Search},
  booktitle     = {ICLR},
  year          = {2018},
  eprint        = {1806.09055},
  archivePrefix = {arXiv},
  publisher     = {OpenReview.net}
}

@inproceedings{lakshminarayanan2017simple,
  title     = {Simple and scalable predictive uncertainty estimation using deep ensembles},
  author    = {Lakshminarayanan, Balaji and Pritzel, Alexander and Blundell, Charles},
  booktitle = {NeurIPS},
  pages     = {6405--6416},
  year      = {2017},
  volume    = {30},
  editor    = {I. Guyon and U. V. Luxburg and S. Bengio and H. Wallach and R. Fergus and S. Vishwanathan and R. Garnett},
  eprint    = {1612.01474},
  archivePrefix = {arXiv}
}

@inproceedings{wen2020neural,
  title     = {Neural Predictor for Neural Architecture Search},
  author    = {Wen, Wei and Liu, Hanxiao and Chen, Yiran and Li, Hai and Bender, Gabriel and Kindermans, Pieter-Jan},
  booktitle = {ECCV},
  pages     = {660--676},
  year      = {2020},
  doi       = {10.1007/978-3-030-58580-8_39},
  editor    = {A. Vedaldi and H. Bischof and T. Brox and J. Frahm},
  publisher = {Springer}
}

@inproceedings{schroff2015facenet,
  title     = {FaceNet: A Unified Embedding for Face Recognition and Clustering},
  author    = {Schroff, Florian and Kalenichenko, Dmitry and Philbin, James},
  booktitle = {CVPR},
  pages     = {815--823},
  year      = {2015},
  doi       = {10.1109/CVPR.2015.7298682}
}

@article{Velickovic2017,
  author        = {Velickovic, Petar and Cucurull, Guillem and Casanova, Arantxa and Romero, Adriana and Li{\`{o}}, Pietro and Bengio, Yoshua},
  title         = {Graph Attention Networks},
  journal       = {CoRR},
  volume        = {abs/1710.10903},
  year          = {2017},
  eprint        = {1710.10903},
  archivePrefix = {arXiv},
  doi           = {10.48550/arxiv.1710.10903}
}

@inproceedings{li2020random,
  title     = {Random Search and Reproducibility for Neural Architecture Search},
  author    = {Li, Liam and Talwalkar, Ameet},
  booktitle = {UAI},
  pages     = {367--377},
  year      = {2020},
  volume    = {124},
  editor    = {J. Peters and D. Sontag},
  publisher = {PMLR}
}

@InProceedings{pmlr-v70-cortes17a,
  title = 	 {{A}da{N}et: Adaptive Structural Learning of Artificial Neural Networks},
  author =       {Corinna Cortes and Xavier Gonzalvo and Vitaly Kuznetsov and Mehryar Mohri and Scott Yang},
  booktitle = 	 {Proceedings of the 34th International Conference on Machine Learning},
  pages = 	 {874--883},
  year = 	 {2017},
  editor = 	 {Precup, Doina and Teh, Yee Whye},
  volume = 	 {70},
  series = 	 {Proceedings of Machine Learning Research},
  month = 	 {06--11 Aug},
  publisher =    {PMLR},
  url = 	 {https://proceedings.mlr.press/v70/cortes17a.html}
}

@InProceedings{Chen_2021_CVPR,
    author    = {Chen, Minghao and Fu, Jianlong and Ling, Haibin},
    title     = {One-Shot Neural Ensemble Architecture Search by Diversity-Guided Search Space Shrinking},
    booktitle = {Proceedings of the IEEE/CVF Conference on Computer Vision and Pattern Recognition (CVPR)},
    month     = {June},
    year      = {2021},
    pages     = {16530-16539}
}

@InProceedings{white2021bananas,
  author    = {White, Colin and Neiswanger, Willie and Savani, Yash},
  title     = {Bananas: Bayesian optimization with neural architectures for neural architecture search},
  booktitle = {Proceedings of the AAAI conference on artificial intelligence},
  year      = {2021},
  volume    = {35},
  number    = {12},
  pages     = {10293--10301},
}

@Article{krizhevsky2009learning,
  author    = {Krizhevsky, Alex and Hinton, Geoffrey and others},
  title     = {Learning multiple layers of features from tiny images},
  year      = {2009},
  publisher = {Toronto, ON, Canada},
}

@Article{xiao2017fashion,
  author  = {Xiao, Han and Rasul, Kashif and Vollgraf, Roland},
  title   = {Fashion-mnist: a novel image dataset for benchmarking machine learning algorithms},
  journal = {arXiv preprint arXiv:1708.07747},
  year    = {2017},
}

@inproceedings{cai2021graphnorm,
  title={Graphnorm: A principled approach to accelerating graph neural network training},
  author={Cai, Tianle and Luo, Shengjie and Xu, Keyulu and He, Di and Liu, Tie-yan and Wang, Liwei},
  booktitle={International Conference on Machine Learning},
  pages={1204--1215},
  year={2021},
  organization={PMLR}
}

@misc{nes_dataset,
  author    = {Alexandr Udeneev and Petr Babkin and Oleg Bakhteev},
  title     = {{NES} Surrogate Architectures},
  year      = {2026},
  publisher = {Hugging Face},
  doi       = {10.57967/hf/8676},
  url       = {https://huggingface.co/datasets/Demoren/nes-surrogate-architectures},
  note      = {Revision b6b3f3c}
}

\appendix

\newpage
\section{NAS-Bench formats}

\begin{figure}[h!]
    \centering
     \includegraphics[width=0.9\linewidth]{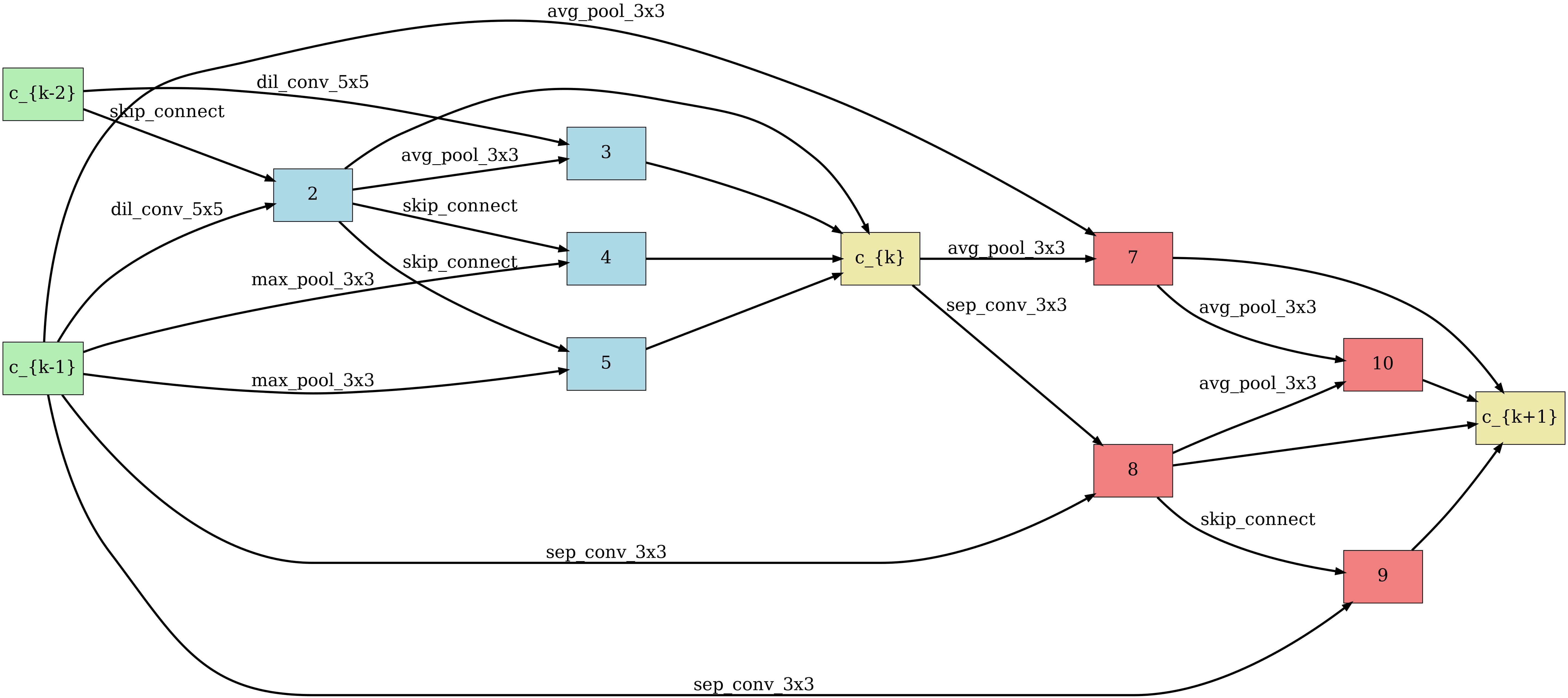}
     \caption{Combined normal and reduced cells. The red vertices belong to the reduction cell; the blue vertices belong to the normal cell.}
     \label{fig:normal_reduction_stacked}
 \end{figure}
    
\newpage
\section{Adversarial attacks}

\begin{figure}[h!]
     \centering
     \includegraphics[width=\linewidth]{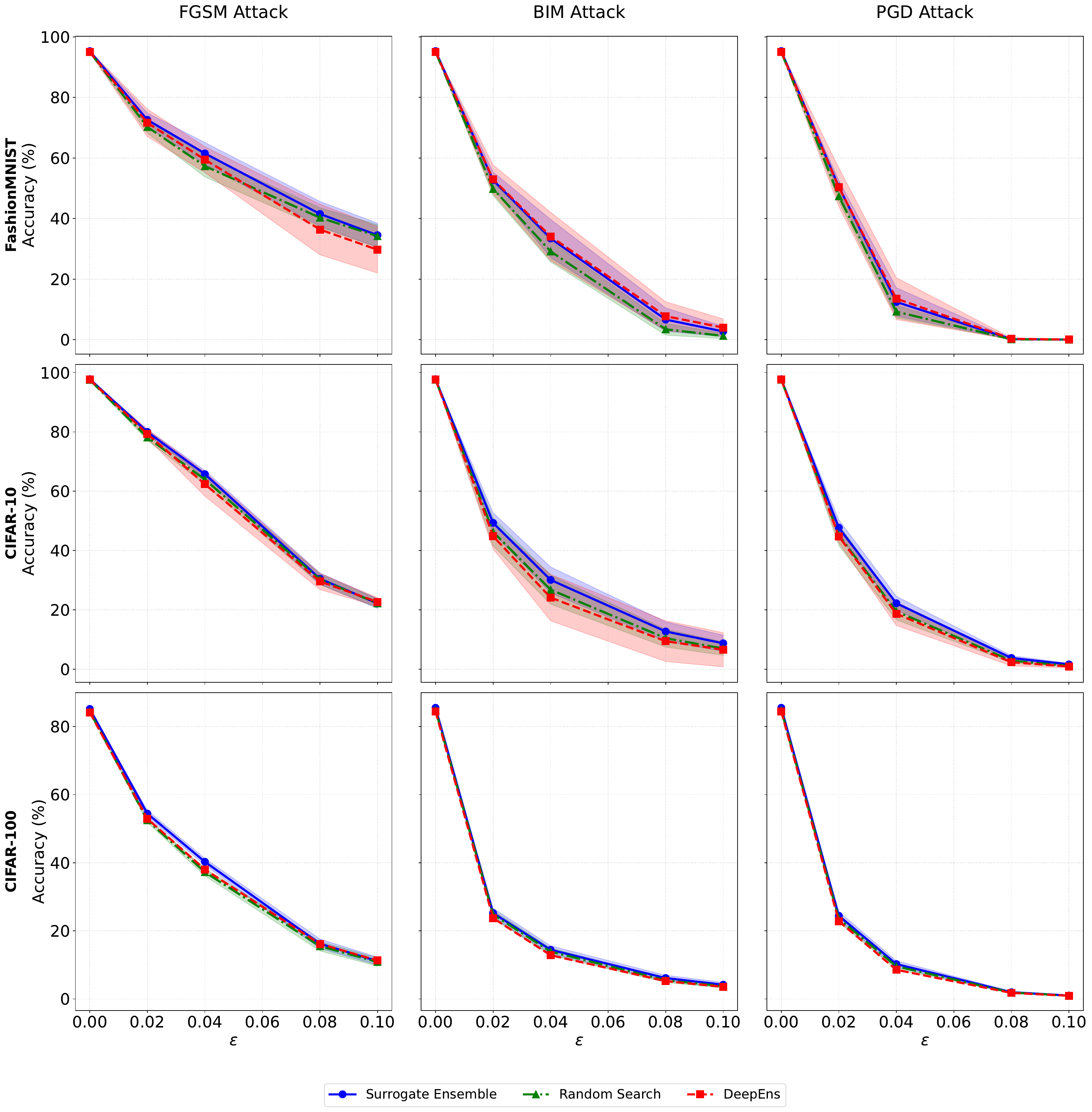}
     \caption{Accuracy of the surrogate ensemble under FGSM, BIM, and PGD attacks across increasing $\varepsilon$ for FashionMNIST, CIFAR-10 and CIFAR-100.}
     \label{fig:adversarial_attacks}
 \end{figure}

Plots of adversarial attacks shown on Figure \ref{fig:adversarial_attacks}.

\newpage
\begin{figure}[h!]
  \centering
     \includegraphics[width=1\linewidth]{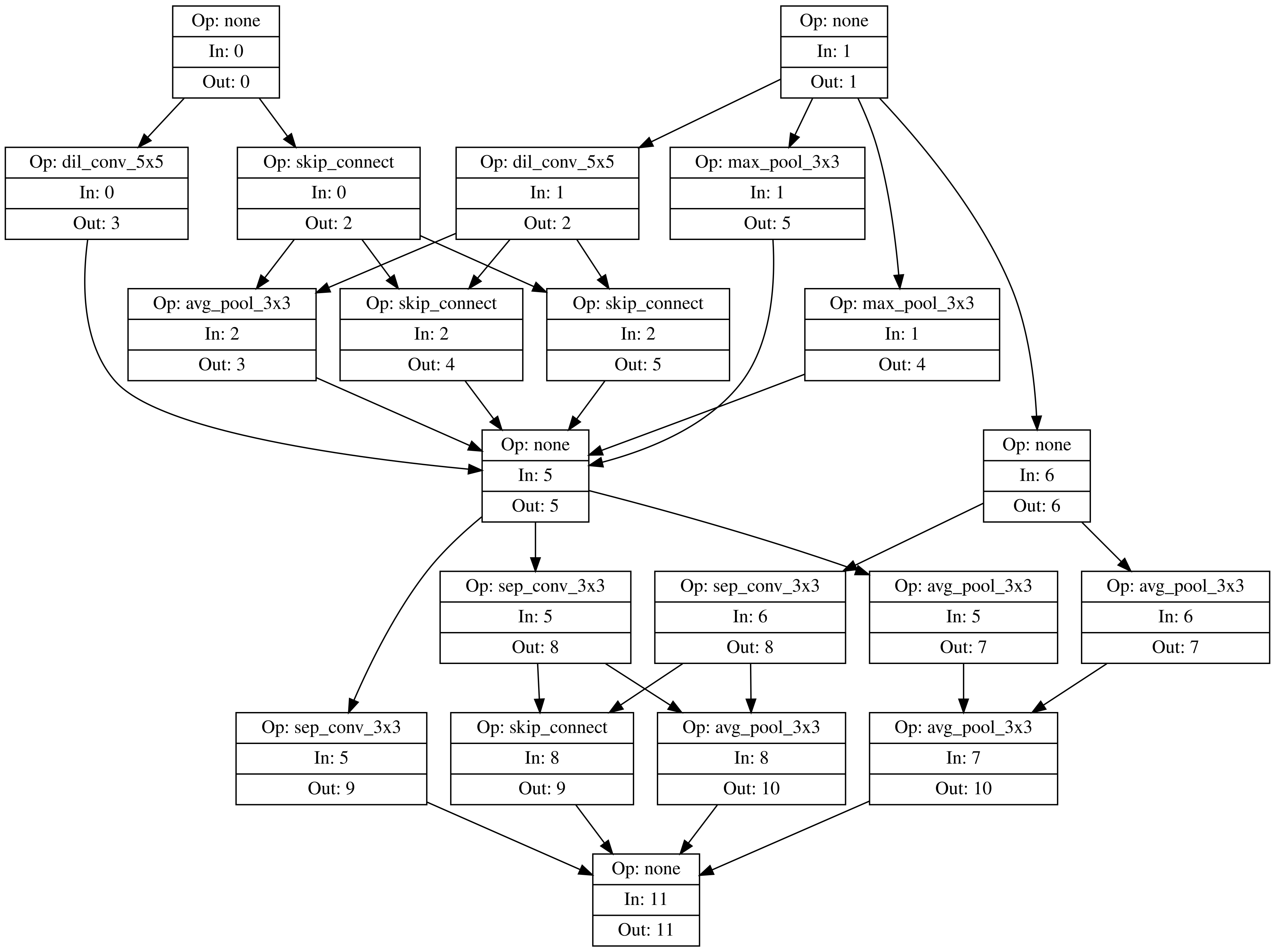}
     \caption{Conversion of an architecture to the NAS-Bench-101 format.}
     \label{fig:NAS-Bench-101}
 \end{figure}

\section{Distributions of models in dataset}

The main distributions of trained models you can see on  Figure \ref{fig:accuracy_distr} and Figure \ref{fig:diversity_distr}:
\newpage
\begin{figure}[h!]
\centering
\begin{minipage}{0.49\textwidth}
\centering
\includegraphics[width=\linewidth]{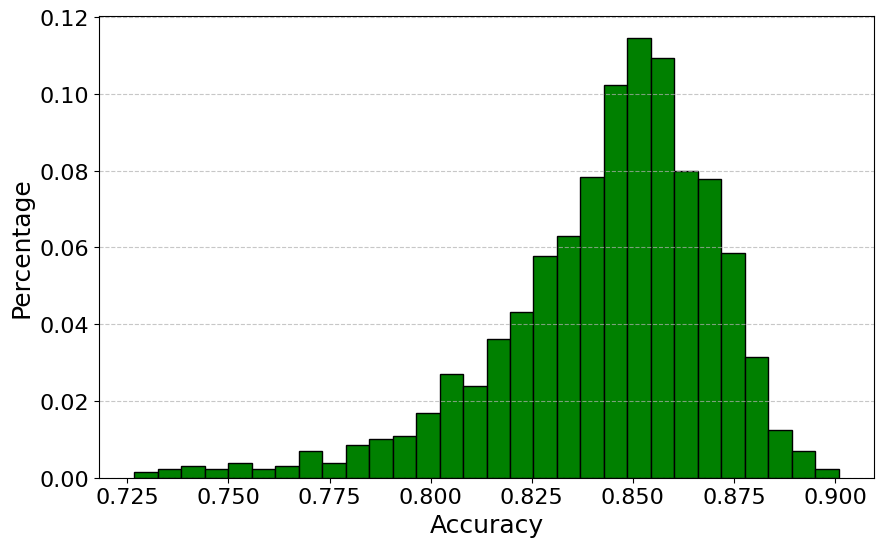}
\caption{Distribution of model accuracies}
\label{fig:accuracy_distr}
\end{minipage}
\hfill
\begin{minipage}{0.49\textwidth}
\centering
\includegraphics[width=\linewidth]{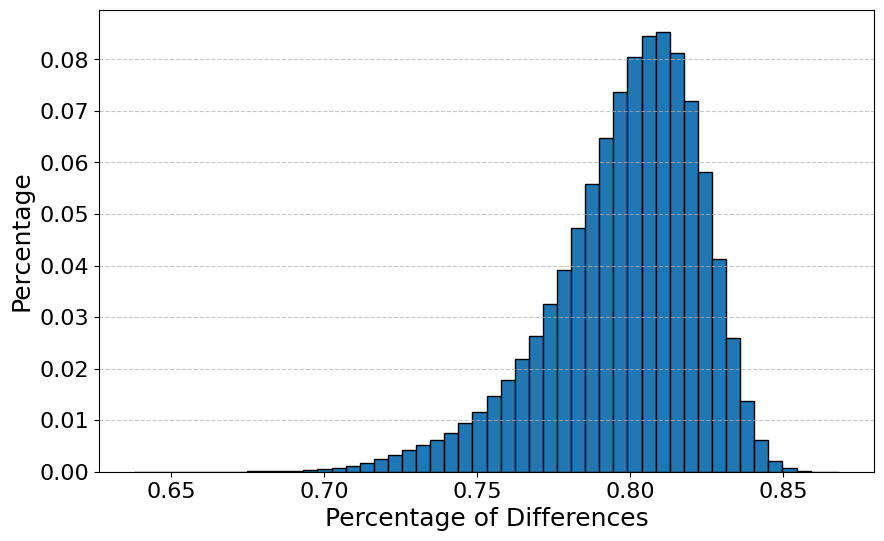}
\caption{Distribution of inter‐model diversity}
\label{fig:diversity_distr}
\end{minipage}
\end{figure}

\section{Training curve for surrogate funcitons}

\begin{figure}[h!]
    \centering
     \begin{minipage}{0.49\textwidth}
         \centering
         \includegraphics[width=\linewidth]{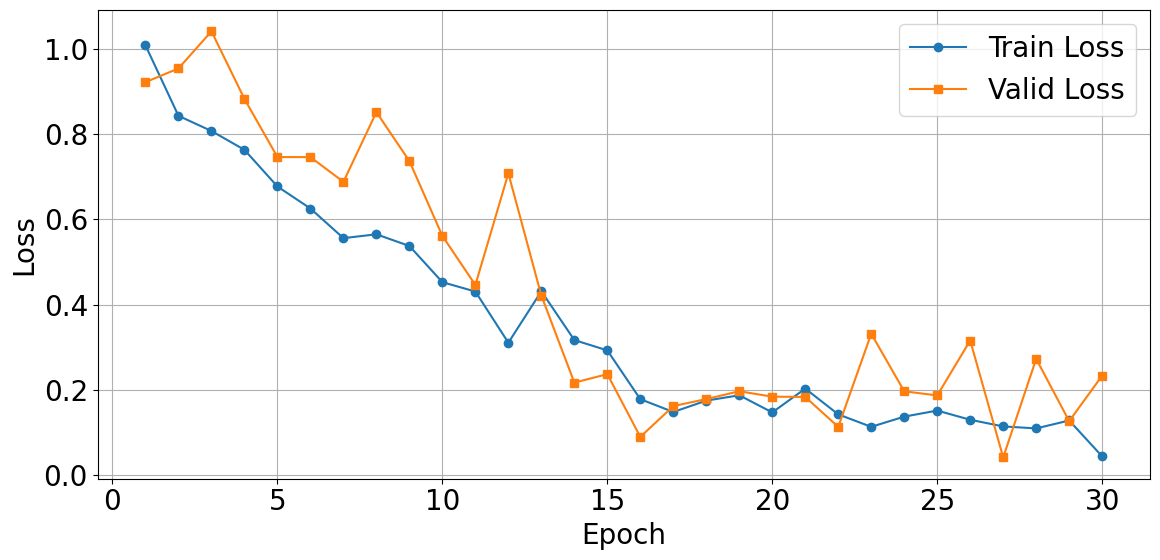}
         \caption{Training curve of the surrogate model for diversity}
         \label{fig:diversity_train}
     \end{minipage}
     \hfill
     \begin{minipage}{0.49\textwidth}
         \centering
         \includegraphics[width=\linewidth]{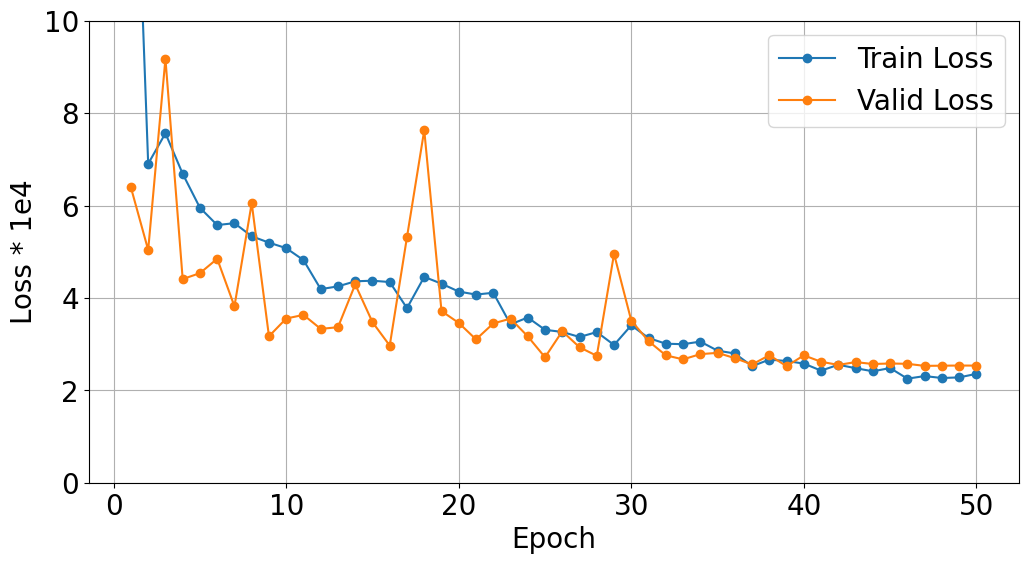}
         \caption{Training curve of the surrogate model for accuracy}
         \label{fig:accuracy_train}
     \end{minipage}
 \end{figure}

You can see the process of training accuracy surrogate function on Figure \ref{fig:accuracy_train} and simularity surrogate function on Figure \ref{fig:diversity_train}

\vspace{2em}

\end{document}